\documentclass[11pt,a4paper]{article}
\usepackage[hyperref]{emnlp-2020}
\usepackage{times}
\usepackage{latexsym}
\usepackage{url}

\usepackage{amsfonts}
\usepackage{amsmath}
\usepackage{booktabs}
\usepackage{multirow}
\usepackage{graphicx}
\usepackage{enumitem}
\usepackage{textcomp}
\usepackage{booktabs}
\usepackage{subcaption}
\usepackage{CJKutf8}

\usepackage[noend]{algpseudocode}
\usepackage{algorithm}

\newcommand{\tabincell}[2]{\begin{tabular}{@{}#1@{}}#2\end{tabular}}  
\newcommand{\chinese}[1]{\begin{CJK*}{UTF8}{gkai}{#1}\end{CJK*}}

\aclfinalcopy % Uncomment this line for the final submission
\newcommand*{\affaddr}[1]{#1}
\newcommand*{\affmark}[1][*]{\textsuperscript{#1}}
\newcommand*{\email}[1]{#1}

\title{Incomplete Utterance Rewriting as Semantic Segmentation}

\author{Qian Liu\affmark[\textdagger]{\thanks{~~Work done during an internship at Microsoft Research.}}~~, Bei Chen\affmark[\S], Jian-Guang Lou\affmark[\S], Bin Zhou\affmark[\textdagger], Dongmei Zhang\affmark[\S]\\
\affaddr{\affmark[\textdagger]School of Computer Science and Engineering, Beihang University, China}\\
\affaddr{\affmark[\S]Microsoft Research, Beijing, China}\\
\affmark[\textdagger]\email{\{qian.liu, zhoubin\}@buaa.edu.cn; \affmark[\S]\{beichen, jlou, dongmeiz\}@microsoft.com}}

\date{}

\begin{document}
\maketitle
\begin{abstract}

Recent years the task of incomplete utterance rewriting has raised a large attention. Previous works usually shape it as a machine translation task and employ sequence to sequence based architecture with copy mechanism. In this paper, we present a novel and extensive approach, which formulates it as a semantic segmentation task. Instead of generating from scratch, such a formulation introduces edit operations and shapes the problem as prediction of a word-level edit matrix. Benefiting from being able to capture both local and global information, our approach achieves state-of-the-art performance on several public datasets. Furthermore, our approach is four times faster than the standard approach in inference.

\end{abstract}

\section{Introduction}

A dramatic progress has been achieved in single-turn dialogue modeling such as open-domain response generation \cite{shang-etal-2015-neural}, question answering \cite{rajpurkar-etal-2016-squad}, etc. By contrast, multi-turn dialogue modeling is still in its infancy, as users tend to use incomplete utterances which usually omit or refer back to entities or concepts appeared in the dialogue context, namely ellipsis and coreference. According to previous studies, ellipsis and coreference exist in more than $70\%$ of the utterances \cite{su-etal-2019-improving}, for which a dialogue system must be equipped with the ability of understanding them. To tackle the problem, early works include learning a hierarchical representation \cite{serban2017hierarchical,zhang-etal-2018-context} and concatenating the dialogue utterances selectively \cite{yan-2016-learning}. Recently, researchers focus on a more explicit and explainable solution: the task of \textbf{I}ncomplete \textbf{U}tterance \textbf{R}ewriting (\textbf{IUR}, also known as context rewriting) \cite{kumar-joshi-2016-non,su-etal-2019-improving,liu-etal-2019-split,pan-etal-2019-improving,elgohary-etal-2019-unpack,zhou-etal-2019-unsupervised}. IUR aims to rewrite an incomplete utterance into an utterance which is semantically equivalent but self-contained to be understood without context. As shown in Table~\ref{tab:example_kobe}, the incomplete utterance $\mathbf{x}_3$ not only omits the subject ``\chinese{北京}''(Beijing), but also refers to the semantic of ``\chinese{阴天}''(cloudy) via ``\chinese{这样}''(this). By explicitly recovering the hidden semantics behind $\mathbf{x}_3$ into $\mathbf{x}_3^*$, IUR makes the downstream dialogue modeling more precise.

\begin{table}[t]
    \centering
    \scalebox{0.9}{
        \begin{tabular}{cc}
            \toprule
            \textbf{Turn} & \textbf{Utterance} (\textit{Translation}) \\
            \hline
            \rule{0pt}{4ex}
            $\mathbf{x}_1$ (A) &  \tabincell{c}{\chinese{北京今天天气如何} \\ \textit{How is the weather in Beijing today} }
            \\ 
            \hline
            \rule{0pt}{4ex}
            $\mathbf{x}_2$ (B) & \tabincell{c}{\chinese{北京今天是阴天} \\ \textit{Beijing is cloudy today} } \\
            \hline
            \rule{0pt}{4ex}
            $\mathbf{x}_3$ (A) & \tabincell{c}{\chinese{为什么总是这样} \\ \textit{Why is always this} } \\
            \hline
            \rule{0pt}{4ex}
            $\mathbf{x}_3^*$ & \tabincell{c}{\chinese{北京为什么总是阴天} \\ \textit{Why is Beijing always cloudy} } \\
            \bottomrule
        \end{tabular}
    }
    \caption{An example dialogue between user A and B, including the context utterances ($\mathbf{x}_1$, $\mathbf{x}_2$), the incomplete utterance ($\mathbf{x}_3$) and the rewritten utterance ($\mathbf{x}_3^*$).}
    \label{tab:example_kobe}
\end{table}
 
To deal with IUR, a natural idea is to transfer models from coreference resolution \cite{clark-manning-2016-improving}. However, this idea is not easy to realize, as ellipsis also accounts for a large proportion. Despite being different, coreference and ellipsis both can be resolved without introducing out-of-dialogue words in most cases. That is to say, words of the rewritten utterance are nearly from either the context utterances or the incomplete utterance. Observing it, most previous works employ the pointer network \cite{oriol-2015-pointer} or the sequence to sequence model with copy mechanism \cite{gu-etal-2016-incorporating,see-etal-2017-get}. However, they generate the rewritten utterance from scratch, neglecting a key trait that the main structure of a rewritten utterance is always the same as the incomplete utterance. To highlight it, we imagine the rewritten utterance as the outcome after a series of edit operations (i.e. substitute and insert) on the incomplete utterance. Taking the example from Table~\ref{tab:example_kobe}, $\mathbf{x}_3^*$ can be obtained by substituting ``\chinese{这样}''(this) in $\mathbf{x}_3$ with ``\chinese{阴天}''(cloudy) in $\mathbf{x}_2$ and inserting ``\chinese{北京}''(Beijing) before ``\chinese{为什么}''(Why), much easier than producing $\mathbf{x}_3^*$ via decoding word by word. These edit operations are carried out between word pairs of the context utterances and the incomplete utterance, analogous to semantic segmentation (a well-known task in computer vision): Given relevance features between word pairs as an image, the model is to predict the edit type for each word pair as a pixel-level mask (elaborated in Section~\ref{sec:method}). Inspired by the above, in this paper, we propose a novel and extensive approach which formulates IUR as semantic segmentation\footnote{Our code is available at \url{https://github.com/microsoft/ContextualSP}.}. Our contributions are as follows:

\begin{itemize}[leftmargin=*]\setlength\itemsep{-0.2em}
    \item As far as we know, we are the first to present such a highly extensive approach which formulates the incomplete utterance rewriting as a semantic segmentation task. 
    \item Benefiting from being able to capture both local and global information, our approach achieves state-of-the-art performance on several datasets across different domains and languages.
    \item Furthermore, our model predicts the edit operations in parallel, and thus obtains a much faster inference speed than traditional methods.
\end{itemize}

\section{Related Work}

The most related work to ours is the line of incomplete utterance rewriting. Recently, it has raised a large attention in several domains. In question answering, previous works include non-sentential utterance resolution using the sequence to sequence based architecture \cite{kumar-joshi-2016-non}, incomplete follow-up question resolution via a retrieval sequence to sequence model \cite{kumar_2017} and sequence to sequence model with a copy mechanism \cite{elgohary-etal-2019-unpack, quan-etal-2019-gecor}. In conversational semantic parsing, \citet{liu_fanda_2019} proposed a novel approach which considers the structures of questions, while \citet{liu-etal-2019-split} imposed an intermediate structure span and decomposed the incomplete utterance rewriting into two sub-tasks. In dialogue generation, \citet{pan-etal-2019-improving} presented a cascaded model which first picks words from the context via BERT, and then combines these words to generate the rewritten utterance, and \citet{su-etal-2019-improving} distinguished the weights of context utterances and the incomplete utterance using a hyper-parameter $\lambda$. Different from all of them, we formulate the task as a semantic segmentation task.

\begin{figure}[t]
    \centering
    \includegraphics[width=0.4\textwidth]{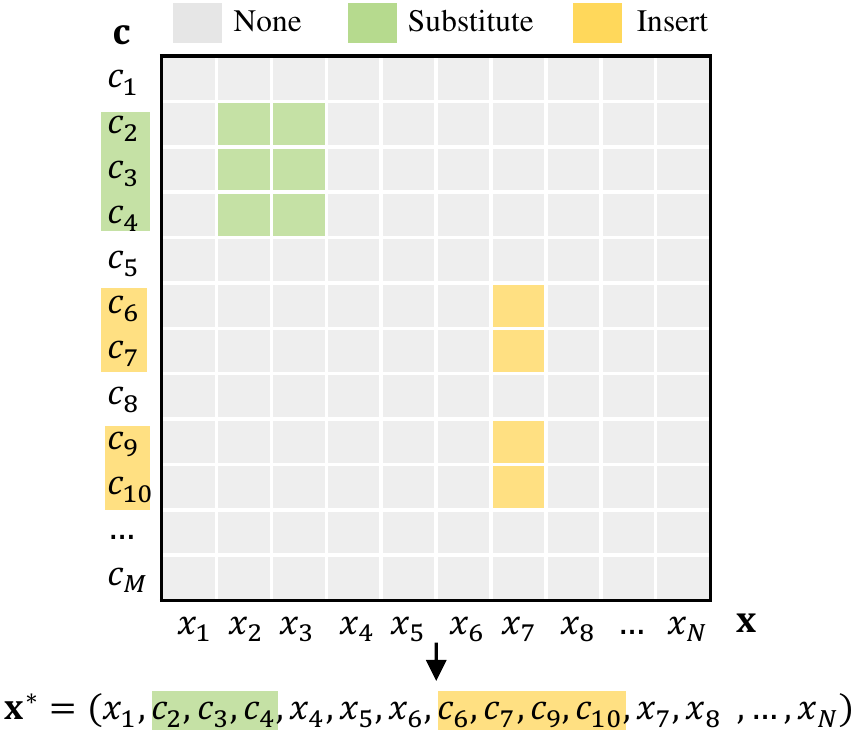}
    \caption{The illustration of the word-level edit matrix applied in our formulation. Each cell belongs to one of three edit types: {\small\texttt{None}}, {\small\texttt{Substitute}} and {\small\texttt{Insert}}.}
    \label{fig:iur-ss}
\end{figure}

Our work is also closely related to coreference resolution. It is an active task that has been studied years, and deep learning based methods have achieved state-of-the-art performance via the paradigm of scoring span or mention pairs \cite{clark-manning-2015-entity,clark-manning-2016-improving,lee-etal-2017-end,lee-etal-2018-higher}. Researchers also explored to use unsupervised contextualized representations to enhance the coreference resolution. \citet{joshi-etal-2019-bert} applied SpanBERT \cite{mandar_span_bert_2019} to enhance the span representation in coreference resolution, and \citet{wu2019coreference} formulated coreference resolution as query-based span prediction and employed SpanBERT to solve it as a machine reading task.
The above works only focus on coreference resolution, while our work deals with coreference and ellipsis under a unified approach.

\begin{figure*}[t]
    \centering
    \includegraphics[width=1.00\textwidth]{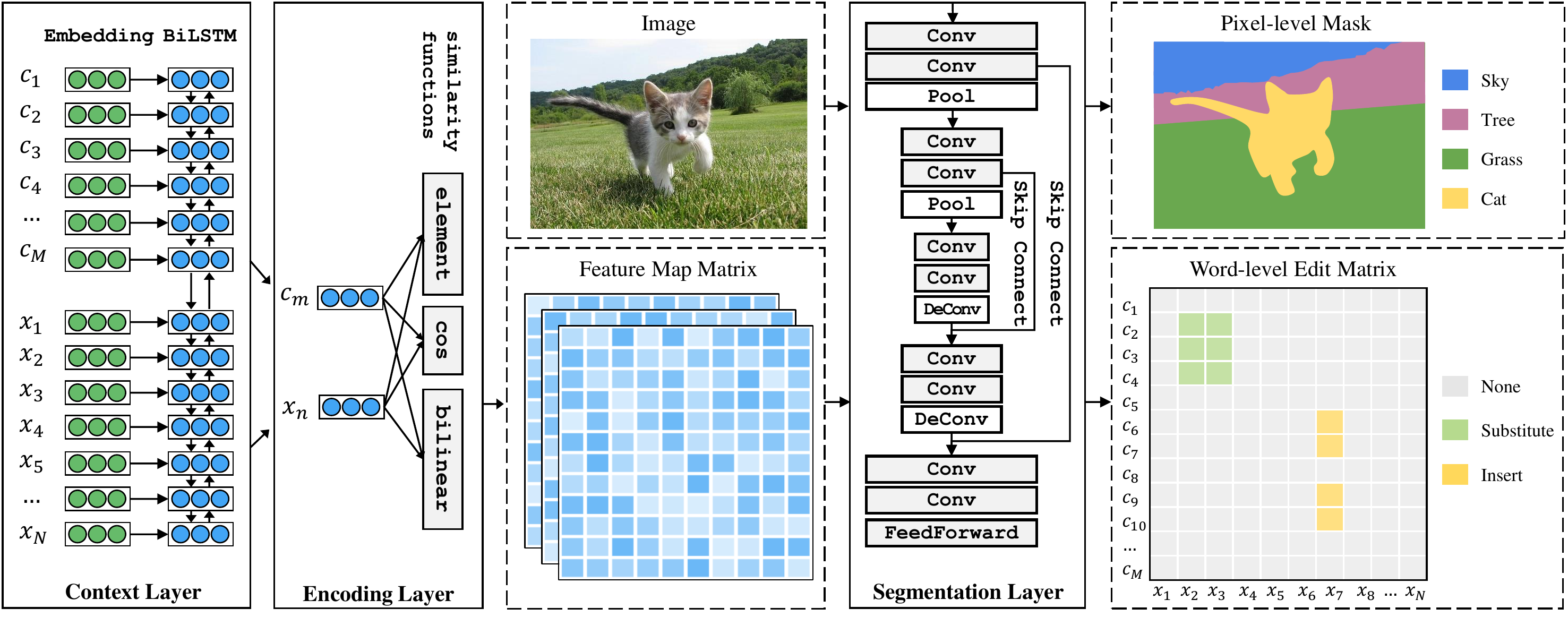}
    \caption{The illustration of the word-level edit matrix construction. The dashed boxes represent the intermediate results of our proposed model (\textbf{bottom}) and their counterparts in semantic segmentation (\textbf{above}). Inside the segmentation layer, a ``Conv'' module consists of a convolutional neural network, batch normalization and an activation function ReLU. ``Pool'' and ``DeConv'' are short for max pooling and deconvolution neural network respectively. }
    \label{fig:model_framework}
\end{figure*}

From the perspective of the methodology, our work is correlated with directions of edit-based text generation and semantic similarity measurement. \citet{wu2019response} proposed a prototype-then-edit paradigm for open-domain response generation, while \citet{malmi-etal-2019-encode} cast text generation as a text editing task and tackled it with a sequence tagging approach. Our work is different from theirs since we model the editing process between two sentences as a semantic segmentation task. As for semantic similarity measurement, similar to us, both \citet{he-lin-2016-pairwise} and \citet{pang2016text} used convolutional neural networks to capture similarities between sentences.

\section{Incomplete Utterance Rewriting as Semantic Segmentation}\label{sec:method}

In this section, we will have a glance at the fundamental idea behind our approach: incomplete utterance rewriting as semantic segmentation.

In a multi-turn dialogue, given the context utterances $(\mathbf{x}_1, \cdots, \mathbf{x}_{t-1})$ and the incomplete utterance $\mathbf{x}_t$, IUR is to rewrite $\mathbf{x}_t$ to $\mathbf{x}^*_t$ using contextual information, where $\mathbf{x}^*_t$ has the same meaning with $\mathbf{x}_t$. The rewritten utterance $\mathbf{x}^*_t$ has self-contained semantics and can be understood solely. To produce $\mathbf{x}^*_t$, our approach formulates the problem as a semantic segmentation task. Concretely, we concatenate all the context utterances to produce an $M$-length word sequence $\mathbf{c}=(c_1,c_2,\cdots,c_M)$. To separate context utterances in different turns, we insert a special word $\mathtt{[S]}$ between each context utterance. Meanwhile, the incomplete utterance is denoted by $\mathbf{x}=(x_1,x_2,\cdots,x_N)$. As mentioned, the rewritten utterance $\mathbf{x}^*$ can be obtained by editing the incomplete utterance $\mathbf{x}$ with in-dialogue words (i.e. words in $\mathbf{c}$). To model edit operations between $\mathbf{x}$ and $\mathbf{c}$, we define a $M\times N$ matrix $Y$, where the entry $Y_{m,n}$ indicates the edit type between $c_m$ and $x_n$. There are three kinds of edit types: {\small\texttt{Substitute}} means replacing the span in $\mathbf{x}$ with the corresponding context span in $\mathbf{c}$; {\small\texttt{Insert}} aims to insert the context span before a certain token in $\mathbf{x}$, and {\small\texttt{None}} represents no operation. For example, as shown schematically in Figure \ref{fig:iur-ss}, we can edit $\mathbf{x}$ by replacing $(x_2,x_3)$ with $(c_2,c_3,c_4)$ and insert $(c_6,c_7,c_9,c_{10})$ before $x_7$. It is notable that we append a special word $\mathtt{[E]}$ to $\mathbf{x}$, to enable {\small\texttt{Insert}} take place after $\mathbf{x}$. More concrete examples can be found in Section \ref{sec:closer_analysis}.

Then, we propose to emit such a matrix $Y$ in a way analogous to the task of semantic segmentation. Specially, we build a $M\times N$ feature map via capturing the word-to-word relevance between $\mathbf{c}$ and $\mathbf{x}$. Taking the feature map as an image, the output word-level edit matrix $Y$ is parallel to the pixel-level mask in semantic segmentation, which bridges IUR and semantic segmentation. Such a formulation comes with several key advantages:
(i) \textbf{Easy}: compared with traditional methods generating the rewritten utterance from scratch, such a formulation introduces edit operations to lower the difficulty of generation;
(ii) \textbf{Fast}: these edits are predicted concurrently, so our model naturally enjoys the fast inference speed than conventional models which decode word by word;
(iii) \textbf{Transferable}: taking the formulation as a bridge between IUR and semantic segmentation, one can transfer empirical models from the community of semantic segmentation with ease.

\section{Methodology}\label{sec:method_model}

As shown in Figure~\ref{fig:model_framework}, our approach firstly obtains the word-level edit matrix through three neural layers. Then based on the word-level edit matrix, it applies a generation algorithm to produce the rewritten utterance. Since the model yields a U-shaped architecture (illustrated later), we name our approach as \textbf{R}ewritten \textbf{U}-shaped \textbf{N}etwork (\textbf{RUN}).

\subsection{Word-level Edit Matrix Construction}

To construct a word-level edit matrix, our model passes through three neural layers: a context layer, an encoding layer and a subsequent segmentation layer. The context layer produces a context-aware representation for each word in both $\mathbf{c}$ and $\mathbf{x}$, based on which the encoding layer forms a feature map matrix $\mathbf{F}$ to capture word-to-word relevance. Finally a segmentation layer is applied to emit the word-level edit matrix.

\paragraph{Context Layer} As shown in the left of Figure~\ref{fig:model_framework}, at first the concatenation of $\mathbf{c}$ and $\mathbf{x}$ passes by the word embedding $\phi$ to get the representation for each word in both utterances. The embedding is initialized using GloVe \cite{pennington-etal-2014-glove}, and then updated along with other parameters. On top of the joint word embedding sequence $\big(\phi(c_1),\!\cdots\!,\phi(c_M),\phi(x_1),\!\cdots\!,\phi(x_N)\big)$, Bidirectional Long Short-Term Memory Network (BiLSTM) \cite{hochreiter1997long,schuster1997bidirectional} is applied to capture contextual information inter and intra utterances.
Although $\mathbf{c}$ and $\mathbf{x}$ are jointly encoded by BiLSTM (see the left of Figure~\ref{fig:model_framework}), below we distinguish their hidden states for clear illustration.
For a word $c_m(m\!=\!1,\dots,M)$ in $\mathbf{c}$, its hidden state is denoted by $\mathbf{u}_m$ obtained through BiLSTM, while the hidden state $\mathbf{h}_n$ is for word $x_n(n\!=\!1,\cdots,N)$ in the incomplete utterance.

\paragraph{Encoding Layer} On top of the context-aware hidden states, we consider several similarity functions to encode the word-to-word relevance. Concretely, for each word $x_n$ in the incomplete utterance and $c_m$ in the context utterances, their relevance is captured by a $D$-dimensional feature vector $\mathbf{F}(x_n, c_m)$. It is produced by concatenating element-wise similarity (Ele Sim.), cosine similarity (Cos Sim.) and learned bi-linear similarity (Bi-Linear Sim.) between them as:
\begin{equation}
    \!\mathbf{F}(x_n, c_m) \!=\! \big[\mathbf{h}_n\!\odot\mathbf{u}_m;\cos({\mathbf{h}_n,\!\mathbf{u}_m)};\mathbf{h}_n\!\mathbf{W}\mathbf{u}_m\big]\!,\!\!\!\!
\end{equation}
where $\mathbf{W}$ is a learned parameter. These similarity functions are expected to model the word-to-word relevance from different perspectives, important for the follow-up edit type classification. 
However, they concentrate on local rather than global information (see discussion in Section~\ref{sec:expr_ablation}). To capture global information, a segmentation layer is proposed.

\paragraph{Segmentation Layer} Taking the feature map matrix $\mathbf{F}\,{\in}\,\mathbb{R}^{M{\times}N{\times}D}$ as a $D$-channel image, the segmentation layer is to predict the word-level edit matrix $Y\,{\in}\,\mathbb{R}^{M{\times}N}$, analogous to a pixel-level mask. Inspired by UNet~\cite{RFB15a}, the layer is formed as a U-shaped structure: two down-sampling blocks and two up-sampling blocks with skip connection. A down-sampling block contains two separate ``Conv'' modules and a subsequent max pooling. Each down-sampling block doubles the number of channels. Intuitively, the down-sampling block expands the receptive fields of each cell, hence providing rich global information for the final decision. An up-sampling block contains two separate ``Conv'' modules, and a subsequent deconvloution neural network. Each up-sampling block halves the number of channels and concatenates the correspondingly cropped feature map in down-sampling as the output (skip connect in Figure~\ref{fig:model_framework}). Finally a feedforward neural network is employed to map each feature vector to one of three edit types, obtaining the word-level edit matrix $Y$.
By incorporating an encoding layer and a segmentation layer, our model is able to capture both local and global information.

\paragraph{BERT Enhanced Embedding} Since pretrained language models have been proven to be effective on several tasks, we also experiment with employing BERT \cite{devlin-etal-2019-bert} to augment our model via BERT enhanced embedding.

\subsection{Rewritten Utterance Generation}

Once a word-level edit matrix is emitted, a subsequent generation algorithm is applied for producing the rewritten utterance. As indicated in Figure~\ref{fig:iur-ss}, to apply edit operations without ambiguity, we assume each edit region in $Y$ is a rectangle. However, the predicted $Y$ is not guaranteed to meet this requirement, indicating the need for a standardization step. Therefore, the overall procedure of generation is divided into two stages: first the algorithm delimits standard edit regions via searching minimal covering rectangles for each connected region; then it manipulates the incomplete utterance based on these standard edit regions to produce the rewritten utterance. Since the second step has been illustrated in Section~\ref{sec:method}, in the following we concentrate on the first standardization step.

In the standardization step, we employ the two-pass algorithm (also known as Hoshen–Kopelman algorithm) to find connected regions \cite{cluster_algorithm_1976}. In a nutshell, the algorithm makes two passes over the word-level edit matrix. The first pass is to assign temporary cluster labels and record equivalences between clusters in an order of left to right and top to down. Concretely, for each cell, if its neighbors (i.e. left or top cells with the same edit type) have been assigned temporary cluster labels, it is labeled as the smallest neighboring label. Meanwhile, its neighboring clusters are recorded as equivalent. Otherwise, a new temporary cluster label is created for the cell. The second pass is to merge temporary cluster labels which are recorded as equivalent. Finally, cells with the same label form a connected region. For each connected region, we use its minimal covering rectangle to serve as the output of our model.

\subsection{Distant Supervision}

As mentioned in Section~\ref{sec:method}, the expected supervision for our model is the word-level edit matrix, but existing datasets only contain rewritten utterances. Therefore, we use a procedure to automatically derive (noisy) word-level edit matrices (i.e. distant supervision), and use these examples to train our model. We use the following process to build our training set. First, we find a \textbf{L}ongest \textbf{C}ommon \textbf{S}ubsequence (\textbf{LCS}) between $\mathbf{x}$ and $\mathbf{x}^*$. Then, for each word in $\mathbf{x}^*$, if it is not in LCS, it is marked as \textsc{Add}. Conversely, for each word in $\mathbf{x}$ but not in LCS, it is marked as \textsc{Del}. Contiguous words with the same mark are merged into one span. By a span-level comparison, any \textsc{Add} span in $\mathbf{x}^*$ with a \textsc{Del} counterpart (i.e. under the same context) relates it to {\small\texttt{Substitute}}. Otherwise, the span is inserted into $\mathbf{x}$, corresponding to {\small\texttt{Insert}}.

Taking the example from Table~\ref{tab:example_kobe}, given $\mathbf{x}$ as ``\chinese{为什么总是这样}''(Why is always this) and $\mathbf{x}^*$ as ``\chinese{北京为什么总是阴天}''(Why is Beijing always cloudy), their longest common subsequence is ``\chinese{为什么总是}''(Why is always). Therefore, with ``\chinese{这样}''(this) in $\mathbf{x}$ being marked as \textsc{Del} and ``\chinese{阴天}''(cloudy) in $\mathbf{x}^*$ being marked as \textsc{Add}, they correspond to the edit type {\small\texttt{Substitute}}. In comparison, since ``\chinese{北京}''(Beijing) cannot find a counterpart, it is related to the edit type {\small\texttt{Insert}}.

\section{Experiments}\label{sec:experiment}

In this section, we conduct thorough experiments to demonstrate the superiority of our approach.

\subsection{Experimental Setup}

\begin{table*}[t]
    \centering
    \scalebox{0.83}{
        \begin{tabular}{lccccccccccccc}
        \toprule
        Model & $\mathcal{P}_1$ & $\mathcal{R}_1$ & $\mathcal{F}_1$ & $\mathcal{P}_2$ & $\mathcal{R}_2$ & $\mathcal{F}_2$ & $\mathcal{P}_3$ & $\mathcal{R}_3$ & $\mathcal{F}_3$ & $\mathbf{B}_1$ & $\mathbf{B}_2$ & $\mathbf{R}_1$ & $\mathbf{R}_2$ \\
        \midrule
        Syntactic\,$^\dagger$ & $67.4$ & $37.2$ & $47.9$ & $53.9$ & $30.3$ & $38.8$ & $\underline{45.3}$ & $25.3$ & $32.5$ & $84.1$ & $81.2$ & $89.3$ & $80.6$ \\
        L-Gen\,$^\dagger$ & $65.5$ & $40.8$ & $50.3$ & $52.2$ & $32.6$ & $40.1$ & $43.6$ & $27.0$ & $33.4$ & $84.9$ & $81.7$ & $88.8$ & $80.3$ \\
        L-Ptr-Gen\,$^\dagger$ & $66.6$ & $40.4$ & $50.3$ & $\underline{54.0}$ & $33.1$ & $41.1$ & $\underline{45.9}$ & $28.1$ & $34.9$ & $84.7$ & $81.7$ & $89.0$ & $80.9$ \\
        RUN (Ours) & $\mathbf{66.9}$ & $\mathbf{54.9}$ & $\mathbf{60.3}$ & $\underline{53.0}$ & $\mathbf{43.4}$ & $\mathbf{47.7}$ & $43.8$ & $\mathbf{35.7}$ & $\mathbf{39.3}$ & $\mathbf{91.1}$ & $\mathbf{88.0}$ & $\mathbf{91.0}$ & $\mathbf{83.3}$ \\
        \midrule
        PAC\,$^\dagger$ & $70.5$ & $58.1$ & $63.7$ & $55.4$ & $45.1$ & $49.7$ & $45.2$ & $36.6$ & $40.4$ & $89.9$ & $86.3$ & $91.6$ & $82.8$ \\
        RUN + BERT (Ours) & $\mathbf{73.2}$ & $\mathbf{64.6}$ & $\mathbf{68.6}$ & $\mathbf{59.5}$ & $\mathbf{53.0}$ & $\mathbf{56.0}$ & $\mathbf{50.7}$ & $\mathbf{45.1}$ & $\mathbf{47.7}$ & $\mathbf{92.3}$ & $ \mathbf{89.6} $ & $\mathbf{92.4}$ & $\mathbf{85.1}$ \\
        \bottomrule
        \end{tabular}
    }
    \caption{The experimental results of (\textbf{Top}) general and (\textbf{Bottom}) BERT-based results on \textsc{Multi}. $\dagger$: Results from \citet{pan-etal-2019-improving}. A bolded \textbf{number} in a column indicates a statistically significant improvement against all the baselines ($p<0.05$), whereas underline \underline{numbers} show comparable performances. Both are same for Table~\ref{tab:rewrite_expr_result}\&\ref{tab:task_canard_result}. }
    \label{tab:multi_expr_result}
\end{table*}

\begin{table}[t]
    \centering
    \scalebox{0.78}{
        \begin{tabular}{lcccc}
        \toprule
        & \textsc{Multi} & \textsc{Rewrite} & \textsc{Task} & \textsc{Canard} \\
        \midrule
        Language & Chinese & Chinese & English & English \\
        \# Ques. (Train) & $194$\,K & $18$\,K & $2.2$\,K & $32$\,K \\
        \# Ques. (Dev) & $5$\,K & $2$\,K & $0.5$\,K & $4$\,K \\
        \# Ques. (Test) & $5$\,K & NA & NA & $6$\,K \\
        Avg. Con len & $25.8$ & $17.7$ & $52.6$ & $85.4$\\
        Avg. Cur len & $8.6$ & $6.5$ & $9.4$ & $7.5$\\
        Avg. Rew len & $12.4$ & $10.5$ & $11.3$ & $11.6$ \\
        \bottomrule
        \end{tabular}
    }
    \caption{Statistics of different datasets. NA means the development set is also the test set. ``Ques'' is short for questions, ``Avg'' for average, ``len'' for length, ``Con'' for context utterance, ``Cur'' for current utterance, and ``Rew'' for rewritten utterance. }
    \label{tab:dataset_statis}
\end{table}

\paragraph{Datasets} We conduct experiments on four public datasets across different domains: Open-Domain Dialogue \big(\textsc{Multi} \citealp{pan-etal-2019-improving}, \textsc{Rewrite} \citealp{su-etal-2019-improving}\big), Task-Oriented Dialogue \big(\textsc{Task} \citealp{quan-etal-2019-gecor}\big) and Question Answering in Context \big(\textsc{Canard} \citealp{elgohary-etal-2019-unpack}\big). We use the same data split for these datasets as their original paper, and some statistics are shown in Table~\ref{tab:dataset_statis}.

\paragraph{Baselines} We consider a bunch of baselines, including LSTM-based models, Transformer-based models and state-of-the-art models on each dataset. (i) \textbf{LSTM-based models} consist of the vanilla sequence to sequence model with attention (L-Gen) \cite{seq2seq_attn_2015}, the pointer network architecture (L-Ptr) \cite{oriol-2015-pointer} and the hybrid pointer generator network (L-Ptr-Gen) \cite{see-etal-2017-get}. (ii) \textbf{Transformer-based models} consist of the basic transformer model (T-Gen) \cite{transformer_2017}, the transformer-based pointer network (T-Ptr), and the transformer-based pointer generator (T-Ptr-Gen). (iii) \textbf{State-of-the-art models} consist of Syntactic \cite{kumar-joshi-2016-non}, PAC \cite{pan-etal-2019-improving}, GECOR \cite{quan-etal-2019-gecor}, L-Ptr-$\lambda$ and T-Ptr-$\lambda$ \cite{su-etal-2019-improving}. We refer readers to their papers for more details. It is remarkable that above methods all generate rewritten utterances from scratch.

\paragraph{Evaluation} We employ both automatic metrics and human evaluations to evaluate our approach. As in literature \cite{pan-etal-2019-improving}, we examine RUN using the widely used automatic metrics BLEU, ROUGE, EM and Rewriting F-score. (i) $\textbf{BLEU}_{n}$ ($\mathbf{B}_{n}$) evaluates how similar the rewritten utterances are to the golden ones via the cumulative $n$-gram BLEU score \cite{papineni-etal-2002-bleu}. (ii) $\textbf{ROUGE}_{n}$ ($\mathbf{R}_{n}$) measures the $n$-gram overlapping between the rewritten utterances and the golden ones, while $\textbf{ROUGE}_{L}$ ($\mathbf{R}_{L}$) measures the longest matching sequence between them \cite{lin-2004-rouge}. (iii) \textbf{EM} stands for the exact match accuracy, which is the strictest evaluation metric. (iv) \textbf{Rewriting} $\textbf{Precision}_{n}$, $\textbf{Recall}_{n}$ and $\textbf{F-score}_{n}$ ($\mathcal{P}_n,\mathcal{R}_n,\mathcal{F}_n$) emphasize more on words from $\mathbf{c}$ which are argued to be harder to copy \cite{pan-etal-2019-improving}. Therefore, they are calculated on the collection of $n$-grams that contain at least one word from $\mathbf{c}$. As validated by \citet{pan-etal-2019-improving}, above automatic metrics are credible indicators to reflect the rewrite quality. However, none of automatic metrics reflects the utterance fluency or the improvement on downstream tasks. Therefore, human evaluations are included to evaluate the fluency of rewritten utterances and their boost on the downstream task.

\paragraph{Implementation Details} Our implementation was based on PyTorch \cite{paszke2017automatic}, AllenNLP \cite{gardner-etal-2018-allennlp} and HuggingFace's transformers library \cite{Wolf2019HuggingFacesTS}. Since the distribution of edit types is severely unbalanced (e.g. {\small\texttt{None}} accounts for nearly $90\%$), we employed weighted cross-entropy loss and tuned the weight on development sets. We used Adam \cite{adam_kingma_2014} to optimize our model and set the learning rate as 1e-3, except for BERT as 1e-5. The embedding size and hidden size in BiLSTM are $100$ and $200$ respectively.
Specifically, BERT mentioned above refer to $\text{BERT}_{\rm base}$. All results of baselines without specific marks were reproduced by ours using OpenNMT with beam size as $4$ \cite{klein-etal-2017-opennmt}.

\begin{table}[t]
    \centering
    \scalebox{0.78}{
        \begin{tabular}{lccccc}
        \toprule
        \textbf{Model} & $\mathbf{EM}$ & $\mathbf{B}_{2}$ & $\mathbf{B}_{4}$ &  $\mathbf{R}_{2}$ & $\mathbf{R}_{L}$ \\
        \midrule
        L-Gen & $47.3$ & $81.2$ & $73.6$ & $80.9$ & $86.3$\\
        L-Ptr-Gen  & $50.5$ & $82.9$ & $75.4$ & $83.8$ & $87.8$\\
        L-Ptr-Net  & $51.5$ &  $82.7$ & $75.5$ & $84.0$ & $88.2$ \\
        L-Ptr-$\lambda$\,$^\dagger$ & $42.3$ & $82.9$ & $73.8$ & $81.1$ & $84.1$ \\
        T-Gen  & $35.4$ & $72.7$ & $62.5$ & $74.5$ & $82.9$ \\
        T-Ptr-Gen  & $53.1$ & $84.4$ &  $77.6$ & $\underline{85.0}$ & $89.1$ \\
        T-Ptr-Net  & $53.0$ & $83.9$ &  $77.1$ & $\underline{85.1}$ & $88.7$ \\
        T-Ptr-$\lambda$ & $52.6$ & $85.6$ & $78.1$ & $\underline{85.0}$ & $89.0$ \\
        RUN (Ours) & $\mathbf{53.8}$ & $\mathbf{86.1}$ & $\mathbf{79.4}$ & $\underline{85.1}$  & $\mathbf{89.5}$ \\
        \midrule
        T-Ptr-$\lambda$ + BERT & $57.5$ & $86.5$ & $79.9$ & $86.9$ & $90.5$ \\
        RUN + BERT (Ours) & $\mathbf{66.4}$ & $\mathbf{91.4}$ & $\mathbf{86.2}$ & $\mathbf{90.4}$ & $\mathbf{93.5}$ \\
        \bottomrule
        \end{tabular}
    }
    \caption{The experimental results on \textsc{Rewrite}. $\dagger$: Reproduced from the code released by \citet{su-etal-2019-improving}.}
    \label{tab:rewrite_expr_result}
\end{table}

\begin{table*}
    \centering
    \begin{subtable}[t]{0.35\textwidth}\centering
        \scalebox{0.88}{
            \begin{tabular}{lcccc}
            \toprule
            \multicolumn{4}{c}{\textsc{Task}} \\
            \toprule
            \textbf{Model} & \textbf{EM} & $\mathbf{B}_4$  & $\mathcal{F}_1$ \\
            \midrule
            Ellipsis Recovery\,$^\dagger$ & $50.4$ & $74.1$  & $44.1$ \\
            GECOR 1\,$^\dagger$ & $\underline{68.5}$ & $83.9$  & $66.1$ \\
            GECOR 2\,$^\dagger$ & $66.2$ & $83.0$ & $66.2$ \\
            RUN (Ours) & $\underline{69.2}$ & $\mathbf{85.6}$ & $\mathbf{70.6}$ \\
            \bottomrule
            \end{tabular}
        }
    \end{subtable}
    $\quad\quad$
    \begin{subtable}[t]{0.55\textwidth}\centering
        \scalebox{0.88}{
            \begin{tabular}{lcccccc}
            \toprule
            \multicolumn{7}{c}{\textsc{Canard}} \\
            \toprule
            \textbf{Model} & $\mathbf{B}_{1}$ & $\mathbf{B}_{2}$ & $\mathbf{B}_{4}$ & $\mathbf{R}_{1}$ & $\mathbf{R}_{2}$ & $\mathbf{R}_{L}$ \\
            \midrule
            Copy & $52.4$ & $46.7$ & $37.8$ & $72.7$ & $54.9$ & $68.5$ \\
            Pronoun Sub & $60.4$ & $55.3$ & $47.4$ & $73.1$ & $
            \mathbf{63.7}$ & $73.9$\\
            L-Ptr-Gen & $67.2$ & $\underline{60.3}$ & $\underline{50.2}$ & $\underline{78.9}$ & $62.9$ & $\underline{74.9}$\\
            RUN (Ours) & $\mathbf{70.5}$ & $\underline{61.2}$ & $\underline{49.1}$ & $\underline{79.1}$ & $61.2$ & $\underline{74.7}$ \\
            \bottomrule
            \end{tabular}
        }
    \end{subtable}
    \caption{The experimental results on (\textbf{Left}) \textsc{Task} and (\textbf{Right}) \textsc{Canard}. $\dagger$: Results from \citet{quan-etal-2019-gecor}.}
    \label{tab:task_canard_result}
\end{table*}

\begin{table}[t]
    \centering
    \scalebox{0.85}{
        \begin{tabular}{lccc}
        \toprule
        & \textbf{Win} & \textbf{Tie} & \textbf{Loss} \\
        \midrule
        RUN v.s. L-Ptr-$\lambda$ & $41.6~\%$ & $42.4~\%$ & $16.0~\%$ \\
        RUN v.s. T-Ptr-Gen & ${23.6~\%}$ & $56.4~\%$ & $20.0~\%$  \\
        RUN v.s. T-Ptr-$\lambda$ & ${22.6~\%}$ & $57.0~\%$ & $20.4~\%$ \\
        \bottomrule
        \end{tabular}
    }
    \caption{Pairwise human evaluation results about the rewritten utterance fluency on randomly sampled $500$ dialogues from \textsc{Rewrite}. Our approach achieves similar or better fluency compared with top baselines.}
    \label{tab:pairwise_human_evaluation}
\end{table}

\paragraph{Connection Words} Similar to pointer network \cite{oriol-2015-pointer}, RUN is restricted to predict words which have appeared in the dialogue. Although most examples work well under the restriction, there still exist a few cases which rely on certain words to generate fluent utterances. For example, when rewriting possessive pronouns such as ``their'', we usually need an extra word ``of'' to enhance the fluency. Such common words, named after \textit{connection words}, improve fluency of the rewritten utterances. In practice, we append a small list of connection words to the tail of $\mathbf{c}$, enabling our model to pick connection words as well. For each dataset, their connection word list is automatically derived from the training data.

\subsection{Model Comparison}

Table~\ref{tab:multi_expr_result} and Table~\ref{tab:rewrite_expr_result} show experimental results of our approach and baselines on \textsc{Multi} and \textsc{Rewrite}. As shown, our approach outperforms all baselines significantly. Taking \textsc{Multi} as an example, our approach exceeds the best baseline L-Ptr-Gen by a large margin, reaching a new state-of-the-art performance on almost all automatic metrics. To illustrate, our approach improves the previous best model by $6.4$ points and $10.0$ points on $\mathbf{B}_1$ and $\mathcal{F}_1$ respectively. Furthermore, our approach leaves a striking impression when augmented with BERT. It not only fully surpasses the best sequence generation baseline with BERT (i.e. T-Ptr-$\lambda$+BERT on \textsc{Rewrite}), but also obtains a considerable boost over a cascade model designed for stimulating potential of BERT (i.e. PAC on \textsc{Multi}). Even for the most challenging metric $\mathbf{EM}$ on \textsc{Rewrite}, RUN with BERT improves $8.9$ points, demonstrating the superiority of our model.
Besides, our approach also achieves comparable or better results against all baselines on \textsc{Task} and \textsc{Canard}, as shown in Table~\ref{tab:task_canard_result}. 

\begin{table}[t]
    \centering
    \scalebox{0.8}{
        \begin{tabular}{lccccc}
        \toprule
         & Origin & L-Gen &  L-Ptr-Gen & RUN & Gold \\
        \midrule
        \textbf{Avg. Score} & $0.92$ & $0.93$ & $0.91$ & $1.09$ & $1.10$ \\
        \textbf{NR} & $100\%$ & $74\%$ & $68\%$ & $51\%$ & $46\%$ \\
        \bottomrule
        \end{tabular}
    }
    \caption{Human rating evaluations about the response quality on sampled $300$ dialogues from the development set of \textsc{Multi}. The score ranges from $0$ to $2$. ``NR'' represents the proportion of rewritten utterances which are equal to current utterances.}
    \label{tab:human_eval_response_quality}
\end{table}

Besides automatic results, we perform two groups of human evaluation to answer (i) how fluent the rewritten utterances are and (ii) how much IUR can contribute to downstream tasks. For the evaluation of fluency, we randomly sampled $500$ dialogues in the development set of \textsc{Rewrite}. Then we fed them to representative IUR models and presented generated rewritten utterances to $10$ judges, who are asked to decide which of the rewritten utterances is of higher fluency in pairwise comparisons. Ties are acceptable. Table~\ref{tab:pairwise_human_evaluation} shows the evaluation results. In comparison to the best baseline T-Ptr-$\lambda$, our model only loses in $20.4\%$ cases, which is extremely competitive.

\begin{table}[t]
    \centering
    \scalebox{0.78}{
        \begin{tabular}{clcrcc}
        \toprule
        \textbf{Beam} & \textbf{Model} & $\mathbf{B}_4$ & \multicolumn{1}{c}{$\Delta\mathbf{B}_4$}& \textbf{Latency} & \textbf{Speedup} \\
        \midrule
        \multirow{6}{*}{4} & L-Gen & $73.6$ & $0.0$ & ~~$82$ ms & $1.00~\times$\\
        & L-Ptr-Net & $75.5$ & +$1.9$ & $116$ ms & $0.71~\times$\\
        & L-Ptr-Gen & $75.4$ & +$1.8$ & $110$ ms & $0.75~\times$\\
        & T-Gen & $62.5$ & -$11.1$ & $322$ ms & $0.25~\times$\\
        & T-Ptr-Net & $77.1$ & +$3.5$ & $576$ ms & $0.14~\times$ \\
        & T-Ptr-Gen  & $77.6$ & +$4.0$ & $415$ ms & $0.20~\times$\\
        \midrule
        \multirow{6}{*}{1} & L-Gen & $73.5$ & -$0.1$ & ~~$55$ ms & $1.49~\times$\\
        & L-Ptr-Net & $76.2$ & +$3.0$ & ~~$95$ ms & $0.86~\times$ \\
        & L-Ptr-Gen & $73.3$ & -$0.3$ & ~~$59$ ms & $1.39~\times$\\
        & T-Gen & $60.9$ & -$12.7$ & $240$ ms & $0.38~\times$\\
        & T-Ptr-Net & $77.9$ & +$4.3$ & $401$ ms & $0.20~\times$ \\
        & T-Ptr-Gen & $77.1$ & +$3.5$ & $374$ ms & $0.22~\times$\\
        \midrule
        - & RUN (Ours) & $79.4$ & +$5.8$ & ~~$21$ ms & $3.90 \times$ \\
        \bottomrule
        \end{tabular}
    }
    \caption{The inference speed comparison between RUN and baselines. Beam stands for the beam size in beam search, not applicable for RUN. Latency is computed as the time to produce a single sentence without data batching, averaged over the development set of \textsc{Rewrite}. All models are implemented in PyTorch on a single NVIDIA V100.}
    \label{tab:inferece_speed}
\end{table}

To access the influence of IUR on downstream tasks, we choose multi-turn response selection as a representative, which aims to retrieve suitable responses from a candidate pool considering the context. Concretely, an SMN model trained on the Douban Conversation Corpus is selected as the backbone in multi-turn response selection \cite{wu-etal-2017-sequential}. At first we sampled $300$ dialogues from the development set of \textsc{Multi} as the input to IUR models. Then their predicted rewritten utterances and the context utterances were fed into the SMN model, to help it select suitable responses. The response candidate pool was formed by all utterances in \textsc{Multi}. Finally, $5$ workers were asked to evaluate responses following a multi-scale rating from $0$ to $2$: $0$ means the response is not related to the dialogue; $1$ means the response is related but not interesting enough; and $2$ means the response is satisfying. To illustrate more clearly, we also conduct human rating evaluation on responses under the settings of original dialogue (i.e. without rewriting, relying on the SMN model itself to understand the context) and gold dialogue (i.e. human rewriting). As shown in Table~\ref{tab:human_eval_response_quality}, our model achieves the highest response quality score among IUR models, improving the original setting by $19\%$ relatively. Considering that the SMN model is capable of aggregating implicit context information, it is non-trivial for our model to further improve the response quality.

\begin{table}[t]
    \centering
    \scalebox{0.86}{
        \begin{tabular}{lccccc}
        \toprule
        \textbf{Variant} & $\mathcal{F}_{1}$ & $\mathcal{F}_{2}$ & $\mathcal{F}_{3}$ &  $\mathbf{B}_{2}$ & $\mathbf{R}_{2}$ \\
        \midrule
        RUN & $60.3$ & $47.8$ & $39.4$ & $87.9$ & $83.2$\\
        \midrule
        w/o Edit & ~~$0.0$ & ~~$0.0$ & ~~$0.0$ & $77.4$ & $75.6$ \\
        w/o U-shape Seg. & $55.2$ & $41.4$ & $33.1$ & $86.1$ & $82.5$ \\
        w/o Ele Sim. & $60.4$ & $47.1$ & $38.3$ & $86.8$ & $82.6$ \\
        w/o Cos Sim. & $62.3$ & $48.4$ & $39.2$ & $85.3$ & $82.3$ \\
        w/o Bi-Linear Sim. & $61.6$ & $48.0$ & $39.0$ & $85.8$ & $82.6$\\
        \bottomrule
        \end{tabular}
    }
    \caption{The ablation results on the development set of \textsc{Multi}. ``w/o Edit'' means directly using the current utterance as the rewritten utterance. ``w/o U-shape seg.'' means that our segmentation layer is replaced by a feed-forward neural network with comparable parameters. The remaining variants ablate different similarity functions in the encoding layer.}
    \label{tab:ablation_result}
\end{table}

\begin{figure}
    \centering
    \includegraphics[width=.48\textwidth]{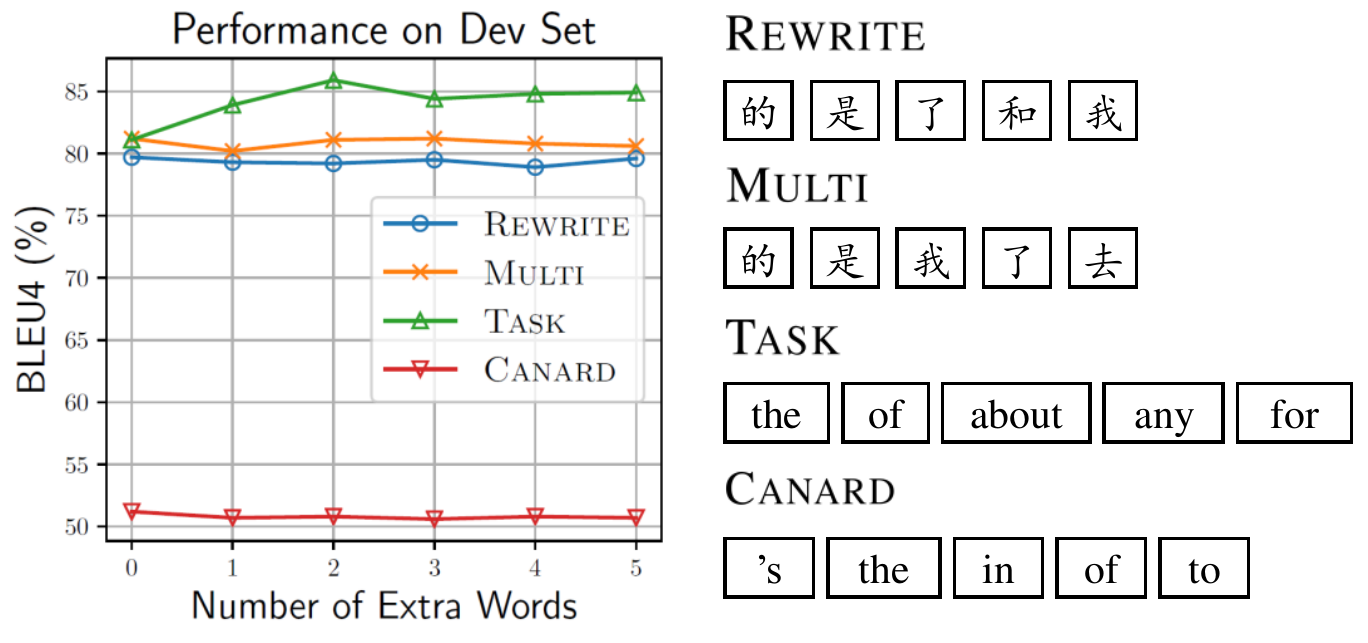}
    \caption{(\textbf{Left}) BLEU4 ($\mathbf{B}_4$) performance with different number of connection words on the development sets of different datasets. (\textbf{Right}) Connection words in decreasing order of frequency on each dataset.}
    \label{fig:ablation_study}
\end{figure}

\subsection{Closer Analysis}\label{sec:closer_analysis}

We conduct a series of experiments to analyze our model deeply. First we conduct an inference speed comparison between our model and representative baselines under the same run-time environment. Then we verify the effectiveness of components in our model by a thorough ablation study. Meanwhile, we touch how the amount of connection words affect the performance. Finally, we present two real cases to illustrate our model concretely.

\begin{figure}[t]
    \centering
    \includegraphics[width=0.45\textwidth]{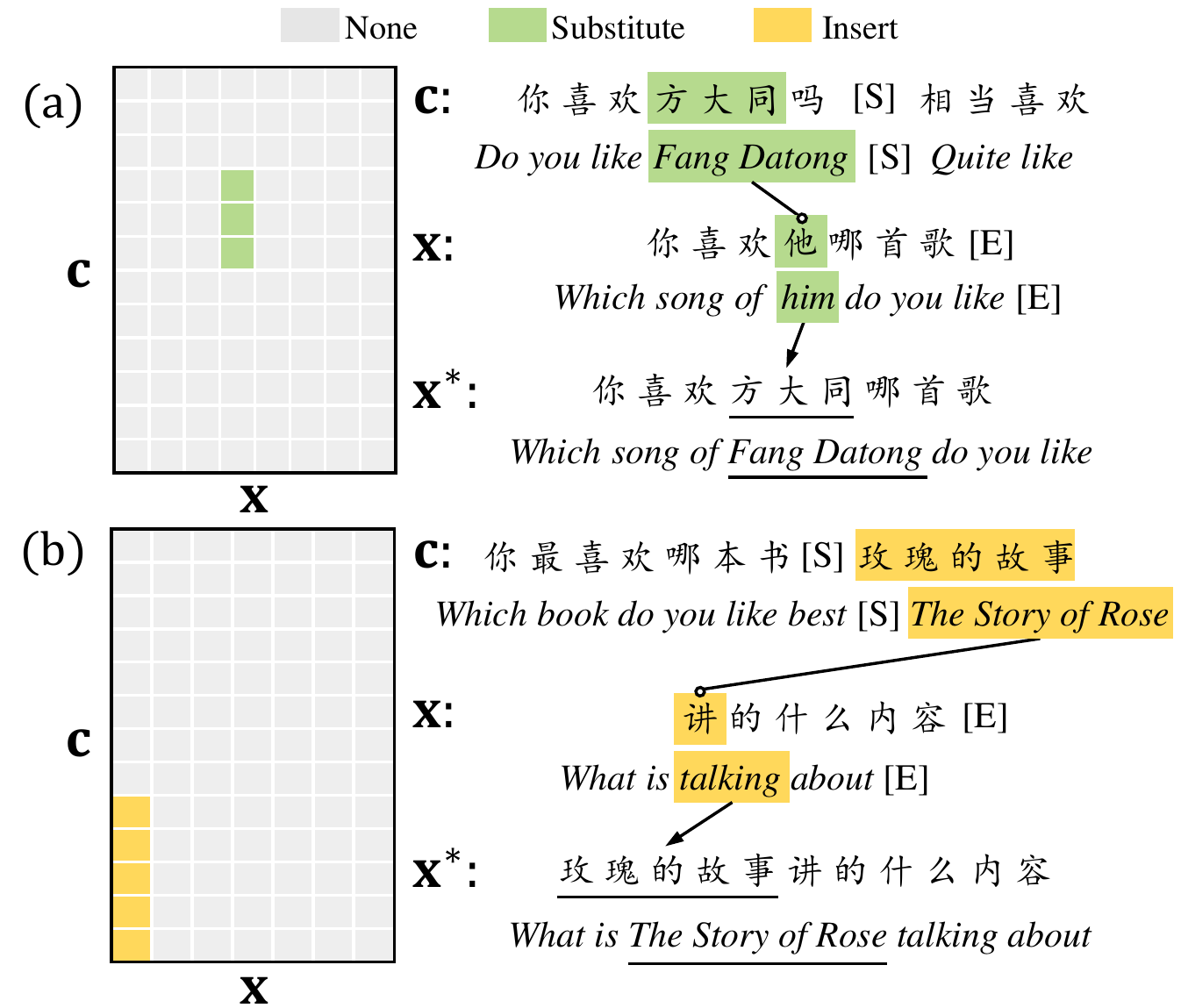}
    \caption{The illustration of (\textbf{Left}) the word-level edit matrix and (\textbf{Right}) the rewritten utterance generation process of two real cases (a) and (b) from \textsc{Rewrite}. }
    \label{fig:case_study}
\end{figure}

\paragraph{Inference Speed} Table~\ref{tab:inferece_speed} compares inference speed between our model and baselines. Since L-Ptr-$\lambda$ and T-Ptr-$\lambda$ are not implemented under PyTorch, we do not show their inference time for fair consideration. Noticing the beam size would affect the inference time of baselines, we also show the results with beam size as 1. Using the simplest L-Gen as a standard, one can find that our model is nearly four times faster, with the highest improvement $\Delta\mathbf{B}_4$. Meanwhile, our model is the only one which can improve both performance and inference speed, significantly surpassing all baselines.

\paragraph{Ablation Study}\label{sec:expr_ablation} 

To verify the effectiveness of different components in our model, we present a thorough ablation study in Table~\ref{tab:ablation_result}. As expected, ``w/o Edit'' causes a huge drop on all evaluation metrics. Notably, the extreme drop on $\mathcal{F}_{n}$ indicates that it is more suited for IUR than common metrics. ``w/o U-shape Seg.'', which ablates the segmentation layer, also brings a great performance drop. Without our segmentation layer capturing global information, an encoding layer only achieves comparable performance with L-Gen, suggesting there are considerable benefits with bridging IUR and semantic segmentation. We also ablates different feature similarity functions (i.e. ``w/o Ele Sim.'', ``w/o Cos Sim.'' and ``w/o Bi-Linear Sim.'') for an in-depth analysis. As shown in Table~\ref{tab:ablation_result}, ablating each similarity function will hurt most metrics. Meanwhile, our model does not depend on any similarity function severely, showing its robustness. Furthermore, we explore how the amount of connection words affect the performance in Figure~\ref{fig:ablation_study}. As indicated, except \textsc{Task}, the number of connection words affect slightly. Nevertheless, it shows a positive effect overall, providing a way to generate out-of-dialogue words. We present two real cases in Figure~\ref{fig:case_study} from \textsc{Rewrite} to illustrate the rewritten process of our model concretely. For both (a) coreference and (b) ellipsis, our model deals with them flexibly.

\subsection{Discussion}

While our approach has made some progress, it still has several limitations. First, our model severely relies on the word order implied by the dialogue. It makes our model vulnerable to some complex cases (i.e. multiple {\small\texttt{Insert}} corresponds to one position). The second limitation is that we predict edit types of each cell independently, ignoring the relationship between neighboring edit types. It is hopefully resolved by the conditional random field algorithm \cite{anurag_2017_crf}. 

The above limitations may raise concerns about the performance upper bound of our approach. In fact, it is not an issue. On three out of four datasets used in the experiments, more than $85$\% examples could be tackled perfectly by our approach ($87.6$\% in \textsc{Task}, $91.0$\% in \textsc{Rewrite}, $95.3$\% in \textsc{Multi}). The number in \textsc{Canard} is relatively low ($42.5$\%) since human annotators introduce many new words in rewriting. Nevertheless, the BLEU upper bound in \textsc{Canard} could be as high as $72.5$\% with our approach, which is acceptable.

The last point we focus on is why similarities can be good features for determining edits. We think it can be elaborated from two aspects. For coreference, the similarity function is suitable for identifying whether two spans refer to the same entity. For ellipsis, the similarity function is an effective indicator to find matching anchors, which indicate the possible insertion positions.

\section{Conclusion \& Future Work}

In this paper, we present a novel and extensive approach which formulates the incomplete utterance rewriting as a semantic segmentation task. On top of the formulation, we carefully design a U-shaped rewritten network, which outperforms existing baselines significantly on several datasets. In the future, we will investigate on extending our approach to more areas.

\section*{Acknowledgments}

We thank all the anonymous reviewers for their valuable comments. This work was supported in part by National Natural Science Foundation of China (U1736217 and 61932003), and National Key R\&D Program of China (2019YFF0302902).

\bibliography{emnlp-2020}
\bibliographystyle{acl_natbib}
\end{document}